\documentclass{article} 
\usepackage{iclr2024_conference,times}

\usepackage{amsmath,amsfonts,bm}

\def\eqref#1{equation~\ref{#1}}

\def\1{\bm{1}}

\def\vp{{\bm{p}}}
\def\vq{{\bm{q}}}

\def\vz{{\bm{z}}}

\def\mC{{\bm{C}}}

\def\mX{{\bm{X}}}

\DeclareMathAlphabet{\mathsfit}{\encodingdefault}{\sfdefault}{m}{sl}
\SetMathAlphabet{\mathsfit}{bold}{\encodingdefault}{\sfdefault}{bx}{n}

\def\emX{{X}}

\newcommand{\R}{\mathbb{R}}

\usepackage{enumitem}
\usepackage{svg}
\usepackage{mathtools}
\usepackage{pifont}
\usepackage{booktabs} 
\usepackage{threeparttable}
\usepackage{multirow}
\usepackage{multicol}
\usepackage{bbm}
\usepackage{colortbl}
\usepackage{caption}
\usepackage{subcaption}
\usepackage[most]{tcolorbox}
\usepackage[ruled, lined, linesnumbered, commentsnumbered, longend]{algorithm2e}
\usepackage[
    outputdir=./,
    frozencache=true,
    cachedir=minted-cache]{minted}
\usepackage[colorlinks=true,allcolors=blue]{hyperref}%

\newcommand{\cmark}{\ding{51}}%
\newcommand{\xmark}{\ding{55}}%
\newcommand{\red}[1]{\textcolor{black}{#1}}
\newcommand{\blue}[1]{\textcolor{black}{#1}}

\definecolor{kellygreen}{rgb}{0.3, 0.73, 0.09}
\definecolor{apricot}{rgb}{0.972549, 0.6196078, 0.4823529}

\newcommand{\first}[1]{\underline{\textbf{#1}}}
\newcommand{\second}[1]{#1}
\newcommand{\up}[1]{\footnotesize{(\textcolor{kellygreen}{$\bm{\uparrow#1}$}})}
\newcommand{\down}[1]{\footnotesize{(\textcolor{red}{$\bm{\downarrow#1}$}})}

\newcommand{\update}[1]{\textcolor{black}{#1}}

\SetCommentSty{mycommfont}

\usepackage{url}

\usepackage{hyperref}

\title{Towards domain-invariant Self-Supervised Learning with Batch Styles Standardization}

\author{Marin Scalbert \& Maria Vakalopoulou \\
MICS, CentraleSupélec, Université Paris-Saclay\\
Gif-sur-Yvette, France\\
\texttt{\{name.surname\}@centralesupelec.fr}
\And
Florent Couzinié-Devy \\
VitaDX\\
Paris, France \\
\texttt{f.couzinie-devy@vitadx.com} \\
}

\newcommand\eg{\textit{e.g.}}

\iclrfinalcopy 

\begin{document}
\maketitle
\begin{abstract}
    In Self-Supervised Learning (SSL), models are typically pretrained, fine-tuned, and evaluated on the same domains. However, they tend to perform poorly when evaluated on unseen domains, a challenge that Unsupervised Domain Generalization (UDG) seeks to address. Current UDG methods rely on domain labels, which are often challenging to collect, and domain-specific architectures that lack scalability when confronted with numerous domains, making the current methodology impractical and rigid. Inspired by contrastive-based UDG methods that mitigate spurious correlations by restricting comparisons to examples from the same domain, we hypothesize that eliminating style variability within a batch could provide a more convenient and flexible way to reduce spurious correlations without requiring domain labels. To verify this hypothesis, we introduce Batch Styles Standardization (BSS), a relatively simple yet powerful Fourier-based method to standardize the style of images in a batch specifically designed for integration with SSL methods to tackle UDG. Combining BSS with existing SSL methods offers serious advantages over prior UDG methods: (1) It eliminates the need for domain labels or domain-specific network components to enhance domain-invariance in SSL representations, and (2) offers flexibility as BSS can be seamlessly integrated with diverse contrastive-based but also non-contrastive-based SSL methods. Experiments on several UDG datasets demonstrate that it significantly improves downstream task performances on unseen domains, often outperforming or rivaling with UDG methods. Finally, this work clarifies the underlying mechanisms contributing to BSS's effectiveness in improving domain-invariance in SSL representations and performances on unseen domain.
\end{abstract}
\section{Introduction}

\textbf{Motivations.} In recent years, Self-Supervised Learning (SSL), has seen significant growth and success~\citep{chen2020simple,grill2020bootstrap,caron2020unsupervised,caron2021emerging,assran2022masked,bardes2021vicreg,he2022masked}.
However, SSL generally assumes that pretraining, fine-tuning and testing data come from the same domains, an assumption which does not hold true in practice and thereby limits its real-life applications. The distribution shifts between pretraining/fine-tuning domains (sources domains) and testing domains (targets domains) usually lead to poor generalization on testing domains.

Unsupervised Domain Generalization (UDG)~\citep{zhang2022towards}, aims to tackle this issue by evaluating how well fine-tuned SSL models generalize to unseen target domains. In UDG, models are first pretrained on unlabeled data, fine-tuned on labeled data and finally evaluated on data from unseen domains. This work focuses on the \textit{all-correlated} UDG setting which is the most standard and studied one ~\citep{zhang2022towards,harary2022unsupervised,yang2022domain}. In this setting, unlabeled and labeled data come from the same source domains, testing data come from unseen target domains, while all cover the same classes.

Current UDG methods suffer from the same drawbacks: (1) They require domain labels to reinforce domain-invariance in SSL representations, while in practice these labels may be challenging to obtain or even unavailable and (2) they all rely on domain-specific architectures, such as domain-specific negative queues or domain-specific decoders, that lack scalability when confronted to numerous domains. These limitations highlight the need for more practical and flexible UDG methods. 

Taking inspiration from contrastive-based UDG methods that reduce spurious correlations by restraining comparisons to examples from the same domain, we believe that removing style variability within a batch through style standardization may provide a more practical and flexible way to mitigate spurious correlations and achieve domain-invariant SSL representations without requiring any domain labels.
\par
\textbf{Contributions. } 
To investigate the effectiveness of style standardization in mitigating spurious correlations within SSL representations with the aim of proposing more convenient and flexible UDG approaches, we introduce Batch Styles Standardization (BSS). BSS is a simple yet powerful Fourier-based method for standardizing image styles within a batch purposefully designed for integration with existing SSL methods to reinforce domain-invariance. Style standardization is performed by transferring the style of a randomly selected image to all images in the batch. 

Integrated with existing SSL methods, it confers significant advantages over prior UDG works: (1) it reinforces domain-invariance without requiring any domain labels or domain-specific architecture, and (2) it offers simplicity and flexibility, integrating easily with \update{contrastive-based (SimCLR~\citep{chen2020simple}, SWaV~\citep{caron2020unsupervised}) but also non-contrastive-based SSL methods (MSN~\citep{assran2022masked}).}

Experiments conducted on UDG datasets indicate that BSS combined with the different SSL methods yields significant performance gains on unseen domains while outperforming or competing with established UDG methods. Finally, extensive experiments have been conducted to clarify the underlying mechanisms driving BSS’s effectiveness in enhancing domain-invariance in SSL representations and performances on unseen domains.

\section{Related works}

\textbf{Domain Generalization. }
    DG aims to learn a model from multiple source domains with distinct distributions to generalize well to unseen target domains.
    Former DG methods have focused on aligning source features distributions using a large panel of techniques ~\citep{ganin2016domain,kang2019contrastive,li2018domain,li2018deep,M3SDA,scalbert2021multi,zhao2020domain}.
    Recently, the trend has shifted towards improving cross-domain generalization by refining data augmentation strategies. These strategies can be applied at either the image level~\citep{scalbert2022test,xu2021fourier,yang2020fda,zhou2020deep,zhou2020learning} or the feature representation level~\citep{kang2022style,li2021simple,mixstyle}, and can be non-parametric~\citep{xu2021fourier,mixstyle}, trained adversarially during the DG task~\citep{hoffman2018cycada,kang2022style,zhou2020deep,zhou2020learning}, or pretrained beforehand on source domains~\citep{scalbert2022test}. Among these methods, Fourier-based Augmentations (FA)~\citep{xu2021fourier,yang2020fda} stand out as a simple and promising approach to instill domain-invariance into the representations and, thereby, enhance generalization. In this work, the proposed BSS extends FA's style transfer ability to standardize the style of images within a batch so as to strengthen domain-invariance in SSL methods.
 
\textbf{Self-Supervised Learning. }
    SSL has gained a lot of attention for its ability to efficiently pretrain models on abundant unlabeled data and subsequently fine-tune them for downstream tasks with limited labeled data. Contrastive and non-contrastive-based methods have emerged as successful approaches. The former focuses on making representations
    of similar examples (positives) closer while pushing apart representations of
    dissimilar examples (negatives). Similar examples are usually built by generating several augmented views of the same image. These methods operate either at the instance-level~\citep{chen2020simple,chen2020improved,hu2021adco} or cluster-level~\citep{deep_cluster,caron2020unsupervised}. Given their reliance on a large number of negatives, non-contrastive-based methods have attempted to eliminate the use of negative examples but require additional tricks to avoid collapse~\citep{grill2020bootstrap,chen2021exploring,caron2021emerging}. In this work, harnessing FA and the proposed BSS, we extend both contrastive-based (SimCLR, SWaV) and non-contrastive-based (MSN) methods to strengthen domain-invariance and address UDG.

\textbf{Unsupervised Domain Generalization. }
    \update{Contrastive-based UDG methods} (DARLING~\citep{zhang2022towards}, BrAD~\citep{harary2022unsupervised}) improve domain-invariance by ensuring that positive and negative examples share the same domain. This constraint mitigates spurious correlations within SSL representations when repelling negative examples from positive ones. To respect this constraint,  DARLING exploits domain-specific adversarial negative queues while BrAD maintains domain-specific negative queues containing past representations of a momentum encoder. Additionally, BrAD learns image-to-image mappings from the different domains to a shared space and compares representations of raw and projected images. \update{As an alternative, DiMAE~\citep{yang2022domain} and CycleMAE~\citep{yang2022cycle} rely on Masked Auto-Encoder~\citep{he2022masked} (MAE) with domain-specific decoders to solve a cross-domain reconstruction task.}
    However, these methods rely on domain labels and complex domain-specific architectures, limiting scalability and adaptability. Inspired by UDG contrastive methods, we propose removing style variability within a batch and without domain labels to reduce spurious correlations in SSL methods resulting in simpler and more flexible UDG approaches.

\section{Method} 

\subsection{Problem formulation}
In the \textit{all-correlated} UDG setting, an unlabeled dataset, a labeled dataset and a test dataset are provided. Unlabeled and labeled training data are drawn from the same source domains $\mathcal{D}_S$ while testing data are drawn from unseen target domains $\mathcal{D}_T$. All data share the same class labels space $\mathcal{Y}$. The goal of UDG is to pretrain a model on the unlabeled data, fine-tune it on the labeled data and achieve good generalization on the test dataset.

\subsection{Batch Styles Standardization}
\label{subsec:fourier_based_augmentation}

\subsubsection{Preliminaries on Fourier-based Augmentations}
Fourier-based Augmentations~\citep{xu2021fourier,yang2020fda} are motivated by a property of the Fourier transform: phase components tend to retain semantic information while amplitude components the style information such as intensity and textures. Therefore, to make the network prioritize semantics over style, FA randomly alters the amplitudes of images during training.

More formally, given an image $\mX \in \R^{H \times W}$, its Fourier transform $\mathcal{F}(\mX)$ along with the corresponding amplitude $\mathcal{A}(\mX)$ and phase $\mathcal{P}(\mX)$ are computed as follows:
\begin{align}
        \mathcal{F}(\mX)(u, v) &= \displaystyle \sum_{h=1}^{H}\sum_{w=1}^{W}\emX_{h, w}e^{-i2\pi \left ( \dfrac{h}{H}u + \dfrac{w}{W}v \right)} \\
        & = \mathcal{A}(\mX)(u,v) e^{-i\mathcal{P}(\mX)(u,v)}
\end{align}
\begin{align}
        \textrm{where }
        \mathcal{A}(\mX) = \sqrt{  \operatorname{Re}\left (\mathcal{F}(\mX)\right ) ^2 + \operatorname{Im}\left (\mathcal{F}(\mX)\right ) ^2}
        \textrm{\:and\:}
        \mathcal{P}(\mX) = \arctan \left(
                \dfrac{\operatorname{Im}\left (\mathcal{F}(\mX)\right )}{\operatorname{Re}\left (\mathcal{F}(\mX)\right )}
            \right)
\end{align}
The amplitude $\mathcal{A}(\mX)$ is then altered by substituting its low-frequency components with those of a randomly selected image $\mathcal{A}(\mX')$ resulting in the altered amplitude $\hat{\mathcal{A}}(\mX)$:
\begin{equation}
\hat{\mathcal{A}}(\mX)(u, v) =  
    \begin{cases}
      \mathcal{A}(\mX')(u, v) & \textrm{if } u \leq r*H \textrm{ and } v \leq r*W, \\ 
      \mathcal{A}(\mX)(u, v) & \textrm{if } u > r*H \textrm{ and } v > r*W,
    \end{cases}
\end{equation}
The strength of the augmentation is controlled by the hyperparameter $r \sim U(r_{min}, r_{max})$ representing the ratio between the substituted amplitude area and the entire amplitude area, where $r_{min}$ and $r_{max}$ stand for the minimum and maximum possible ratios. Finally, an augmented image $\hat{\mX}$ with the same content as the original image $\mX$ and style as the randomly chosen image $\mX'$ can be built by applying the inverse Fourier transform $\mathcal{F}^{-1}$ onto the altered amplitude $\hat{\mathcal{A}}(\mX)$ and unmodified phase $\mathcal{P}(\mX)$:
\begin{equation}
    \hat{\mX} = \mathcal{F}^{-1}\left (\hat{\mathcal{A}}(\mX)e^{-i\mathcal{P}(\mX)}\right)
\end{equation}

\subsubsection{Extend Fourier-based augmentations for Batch Styles Standardization}
Drawing inspiration from contrastive-based UDG methods, we believe that removing style variability within a batch might reduce spurious correlations in SSL methods without requiring domain labels resulting in simpler and more flexible UDG approaches.
Since FA possess a style transfer-like ability, they can be extended to perform styles standardization/harmonization. Concretely, it can be accomplished by transferring the style of a randomly chosen image within the batch to all other images in that batch. Hence, the proposed method is referred to as Batch Styles Standardization. 

\begin{figure}[ht]
    \centering
    \begin{subfigure}[b]{0.45\textwidth}
        \centering
        \includesvg[width=\linewidth]{imgs/BSS}
        \caption{}
        \label{subfig:BSS}
    \end{subfigure}
    \begin{subfigure}[b]{0.25\textwidth}
        \centering
        \includegraphics[height=4.5cm]{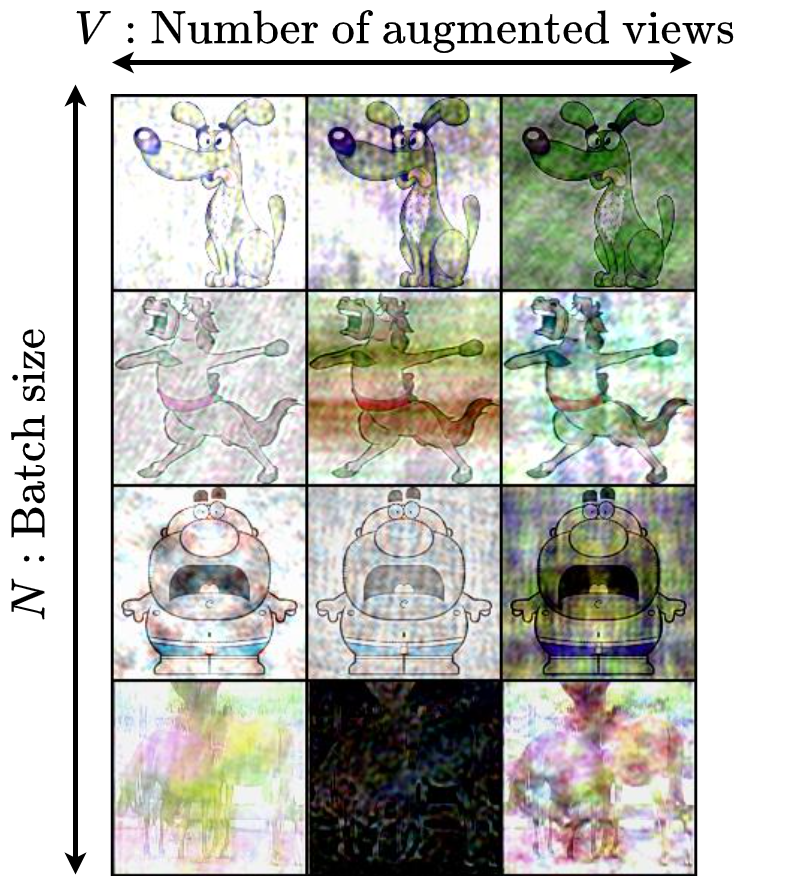}
        \caption{}
        \label{subfig:FA_PACS}
    \end{subfigure}
    \begin{subfigure}[b]{0.25\textwidth}
        \centering
        \includegraphics[height=4.7cm]{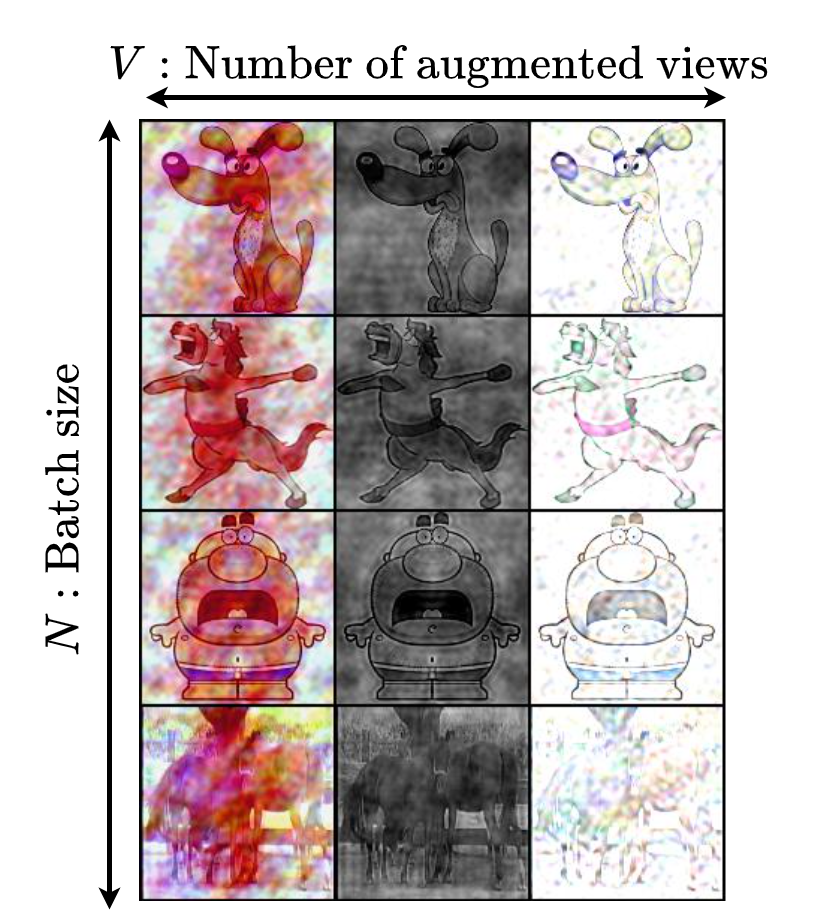}
        \caption{}
        \label{subfig:BSS_PACS}
    \end{subfigure}
    \caption{(a) BSS: Fourier Transform $\mathcal{F}$ is applied on all batch images then low-frequency components of the amplitudes $\mathcal{A}$ (determined by the areas ratio $r$) are replaced by those of a randomly chosen image (the first one in this case). Finally, inverse Fourier transform $\mathcal{F}^{-1}$ is applied to the altered Fourier transforms to build images with standardized styles. (b) Augmented images with standard independent FA. (c) Augmented images with BSS.}
    \label{fig:BSS}
\end{figure}

The process of applying BSS is illustrated on Figure~\ref{subfig:BSS}. Specifically, given a batch of images and their corresponding Fourier transforms, we manipulate the different amplitudes by substituting their low-frequency components with those of a single randomly chosen image. Finally, after applying the inverse Fourier transform to the different modified Fourier transforms, the style of the randomly chosen image is transferred to all images, effectively standardizing/harmonizing the style.
A pseudo-code along with a PyTorch implementation of BSS are provided in
Appendix~\ref{app:pseudo_code}.

To highlight batch-level differences between standard FA and the proposed BSS, we display in Figure~\ref{subfig:FA_PACS} and Figure~\ref{subfig:BSS_PACS}, a $N \times V$ grid of augmented images generated by applying FA or BSS $V$ times on a batch of $N$ images. For a specific view index (column index), it is clear that augmented images produced by FA exhibit different styles whereas in the case of BSS, a unique style prevails. It is important to notice that standardized images can undergo independent geometric augmentations, but color augmentations must be batch-wise to preserve the unique style.

As opposed to UDG methods exploiting domain labels to restrict comparisons to examples from the same domain, standardizing the style of examples using a random style simulates as if they were drawn from the same "pseudo-domain", thereby eliminating the need for domain labels. Moreover, BSS's seamless integration into existing SSL methods removes the need for UDG domain-specific components, such as domain-specific negative queues (DARLING, BRaD) or domain-specific decoders (DiMAE). Collectively, these characteristics position BSS as a simpler and more versatile solution to enhance domain-invariance within SSL methods and address UDG.

\subsection{How to integrate Batch Styles Standardization into SSL methods?}
Both contrastive and non-contrastive methods aim to distribute batch examples over the embedding space. This distribution can be driven by explicit contrastive loss (SimCLR) or methods preventing representation collapse like the Sinkhorn-Knopp algorithm (SWaV, MSN), centering (DINO), or variance regularization (VicReg). However, when dealing with diverse domains/styles within a batch, distributing examples may unintentionally group them by domains/styles, resulting in spurious correlations. To mitigate this, we propose applying BSS to examples undergoing this distribution. Removing style information through standardization should encourage the distribution to focus more on example semantics.
In the following sections, we extend three existing SSL methods (SimCLR, SWaV and MSN) and detail where and how BSS should be integrated. For further technical details about the regular SSL methods, readers may refer to 
Appendix~\ref{app:implementation_details} or the original papers.

\textbf{\textsc{SimCLR}} aims to bring representations of several views of the same image closer (positives) while repelling all other images representations (negatives). In standard SimCLR, each image in a batch of $N$ images is augmented $V$ times resulting in $V$ positive examples. However, this can result in positive examples with domains/styles that differ from those of the negatives (see Figure~\ref{subfig:FA_PACS}), risking unintentional exploitation of this style/domain discrepancy to solve the contrastive task and causing spurious correlations. To address this, we suggest independently applying BSS $V$ times to the initial batch, ensuring that positive and negative examples share the same styles (see Figure~\ref{subfig:BSS_PACS}). 

\textbf{\textsc{SWaV}} computes representations of several views of the same image and clusters them using an online algorithm. Given that representations should capture similar information, SWaV enforces consistency between representations and cluster assignments produced from different views.
To obtain these cluster assignments, the Sinkhorn-Klopp (SK) algorithm~\citep{cuturi2013sinkhorn} is performed on the representations. Concretely, SK solves an optimal transport problem whose constraints are to assign representations to the most similar centroids/prototypes while keeping a uniform assignment distribution over centroids/prototypes. 
However, if several domains/styles are present within views subject to SK, there is a risk of assigning and grouping the corresponding representations using domain/style information resulting in spurious correlations.
To address this, we propose to standardize the style of views subject to SK using BSS. In practice, SWaV employs a multi-crop strategy, generating $2$ global views (large crops) and $V$ local views (small crops) for each image. In this setting, cluster assignments are computed only from the global views while representations are derived from all views. Therefore, we suggest applying BSS only on the global views and augmenting the local views using FA. As this results in $2$ batch of global views, each with its own style, SK is performed on each batch separately.

\textbf{\textsc{MSN}} 
aims to match the representations of masked views of the same image with that of an unmasked view. To derive a view's representation, MSN computes similarities between its embedding and a set of learnable cluster centroids/prototypes and subsequently transforms them into a probability distribution. Since direct matching between masked and unmasked views' representations can lead to representation collapse, MSN simultaneously optimizes a cross-entropy term along with an entropy regularization term on the mean representation of the masked views. This regularization term encourages the model to use the entire set of centroids/prototypes. Additionally, MSN employs SK on the representations of the unmasked views to avoid tuning the hyperparameter weighting the entropy regularization term.
Similarly to SWaV, if several domains/styles are present within unmasked views, assigning the corresponding representations using SK may group examples using domain/style information. To address this, we propose to standardize the style of unmasked views using BSS while we recommend augmenting masked views using FA.

\section{Results}
\label{sec:results}
\subsection{Datasets}
To evaluate the extended SSL methods, experiments were conducted on $3$ datasets commonly used for benchmarking DG / UDG methods, namely \textbf{PACS}, \textbf{DomainNet} and \textbf{Camelyon17 WILDS}.

\textbf{PACS}~\citep{li2017deeper} contains $4$ domains (\textit{photo}, \textit{art painting}, \textit{cartoon}, \textit{sketch}) and 7 classes. \textbf{DomainNet}~\citep{M3SDA} contains $6$ different domains (\textit{clipart}, \textit{infograph}, \textit{quickdraw}, \textit{painting}, \textit{real} and \textit{sketch}) and covers $345$ classes. Following prior UDG works~\citep{harary2022unsupervised,yang2022domain,zhang2022towards}, a subset of DomainNet including $20$ classes out of the $345$ available classes is considered. 
\textbf{Camelyon17 WILDS}~\citep{koh2021wilds} includes images covering $2$ classes (tumor, no tumor) from $5$ domains (hospitals). It is split into \texttt{train}, \texttt{val}, and \texttt{test} subsets comprising respectively $3$, $1$, and $1$ distinct domains.

\subsection{Experimental setup}
Following the standard UDG evaluation protocol~\citep{zhang2022towards}, models were pretrained on source data in an unsupervised way, fine-tuned on a fraction of the source data and finally evaluated on the target data.
For the pretraining step, all our models were pretrained using FA or BSS. We did not use Imagenet~\citep{deng2009Imagenet} transfer learning, except on \textbf{DomainNet} to allow fair comparisons with prior UDG works. This choice is further discussed in
Appendix~\ref{app:transfer_learning_dg_udg}.
For the fine-tuning step, following BrAD, on \textbf{PACS} and \textbf{DomainNet}, all the models were fine-tuned via linear probing except when considering the entire PACS dataset where full fine-tuning was performed like DARLING and DiMAE. On \textbf{Camelyon17 WILDS}, linear probing was performed for each fraction of labeled data.
Pretraining and fine-tuning implementation details are provided in
Appendix~\ref{app:implementation_details}. 

\subsection{Experimental results}
\label{sec:udg_performances}

In the following experiments, the proposed extended SSL methods are compared to regular SSL methods (MoCo V2~\citep{chen2020improved}, SimCLR V2~\citep{chen2020big}, BYOL~\citep{grill2020bootstrap}, AdCo~\citep{hu2021adco}, MAE~\citep{he2022masked}) and UDG methods (DARLING, DiMAE, BRaD, \update{CycleMAE}). For Camelyon17 WILDS, the extended SSL methods are compared to reimplemented UDG methods (DARLING, DiMAE) but also to the Semi-Supervised Learning method FixMatch~\citep{sohn2020fixmatch} and SSL method SWaV trained with additional data from the target.

\textbf{PACS.} For each combination of (sources, target) domains, each fraction of labeled data and each of our SSL models, averaged accuracy over $3$ independent runs are reported on Table~\ref{tab:pacs_udg_performances}. 
Compared to FA, integrating BSS to SimCLR or SWaV, significantly improves the overall accuracy (avg.): SimCLR $\rightarrow (+2.96\%, +4.55\%, +3.61\%, +3.6\%)$; SWaV $\rightarrow (+5.3\%, +5.9\%, +5.46\%, +1.52\%)$ for the fractions of labeled data $1\%$, $5\%$, $10\%$ and $100\%$, respectively.
\update{Extended SSL methods with BSS outperform most of the time other methods, except for the target domain \textit{photo} (BrAD, DiMAE, CycleMAE) or in the $10\%$ (DiMAE) and $100\%$ (DiMAE, CycleMAE) labeled data settings. For the target domain photo, one possible explanation is that other methods benefit from transfer learning on ImageNet while for the $10\%$ and $100\%$ labeled data settings, DiMAE and CycleMAE consider different experimental settings (ViT-base architecture, full fine-tuning on $10\%$.)}
\begingroup
\setlength{\tabcolsep}{3pt}
\begin{table}[]
\caption{
UDG performances on PACS. 
Best methods are highlighted in \first{bold}.}
\label{tab:pacs_udg_performances}
\centering
\begin{threeparttable}[ht]
\resizebox{\linewidth}{!}{%
\begin{tabular}{|l|llll|l|llll|l|}
\hline
 & \multicolumn{5}{c|}{\cellcolor{blue!3}Label Fraction: 1\%} & \multicolumn{5}{c|}{\cellcolor{blue!3}Label Fraction: 5\%} \\
\hline
\multirow{2}{*}{Methods} & \multicolumn{5}{c|}{Target domain} & \multicolumn{5}{c|}{Target domain}\\
\cline{2-11} 
 & \textit{photo} & \textit{art} & \textit{cartoon} & \textit{sketch} & avg. & \textit{photo} & \textit{art} & \textit{cartoon} & \textit{sketch} & avg. \\
\hline
 ERM & 10.90 & 11.21 & 14.33 & 18.83 & 13.82 & 14.15 & 18.67 & 13.37 & 18.34 & 16.13 \\
MoCo V2 & 22.97 & 15.58 & 23.65 & 25.27 & 21.87 & 37.39 & 25.57 & 28.11 & 31.16 & 30.56 \\
SimCLR V2 & 30.94 & 17.43 & 30.16 & 25.20 & 25.93 & 54.67 & 35.92 & 35.31 & 36.84 & 40.68 \\
BYOL & 11.20 & 14.53 & 16.21 & 10.01 & 12.99 & 26.55 & 17.79 & 21.87 & 19.65 & 21.46 \\
AdCo& 26.13 & 17.11 & 22.96 & 23.37 & 22.39 & 37.65 & 28.21 & 28.52 & 30.35 & 31.18 \\
MAE & 30.72 & 23.54 & 20.78 & 24.52 & 24.89 & 32.69 & 24.61 & 27.35 & 30.44 & 28.77 \\
DARLING & 27.78 & 19.82 & 27.51 & 29.54 & 26.16 & 44.61 & 39.25 & 36.41 & 36.53 & 39.20 \\
DiMAE\tnote{*} \  & \second{48.86} & 31.73 & 25.83 & 32.50 & 34.73 & 50.00 & 41.25 & 34.40 & 38.00 & 40.91 \\
BrAD\tnote{*} \ \  & \first{61.81} & \second{33.57} & \second{43.47} & \second{36.37} & \second{43.80} & \first{65.22} & \second{41.35} & \second{50.88} & \second{50.68} & \second{52.03} \\
 CycleMAE\tnote{*} \ \  & 52.63 & 36.25 & 35.53 & 34.85 & 39.82 & 63.24 & 39.96 & 42.15 & 36.35 & 45.43 \\
\hline
\rowcolor{apricot!15!} SimCLR w/ FA & 41.00 & 37.94 & 45.38 & 43.47 & 41.95 & 57.17 & 44.78 & 50.16 & 55.32 & 51.85 \\
\rowcolor{apricot!30!} SimCLR w/ BSS & 43.31 \up{2.31} & \first{38.96} \up{1.02} & \first{48.61} \up{3.23} & \first{48.76} \up{5.29} & \first{44.91} \up{2.96} & \second{58.16} \up{0.99} & \first{46.37} \up{1.59} & \first{55.69} \up{5.53} & \first{65.63} \up{10.04} & \first{56.40} \up{4.55} \\
\hline
\rowcolor{apricot!15!} SWaV w/ FA & 36.15 & 32.93 & 36.63 & 27.37 & 33.27 & 41.64 & 40.95 & 48.51 & 45.32 & 44.10 \\
\rowcolor{apricot!30!} SWaV w/ BSS & 39.74 \up{3.59} & 35.82 \up{2.89} & 42.59 \up{5.96} & 36.12 \up{8.75} & 38.57 \up{5.3} & 50.58 \up{8.94} & 43.00 \up{2.05} & 53.81 \up{5.3} & 52.61 \up{7.29} & 50.00 \up{5.9} \\
\hline
\hline
 & \multicolumn{5}{c|}{\cellcolor{blue!3}Label Fraction: 10\%} & \multicolumn{5}{c|}{\cellcolor{blue!3}Label Fraction: 100\%} \\
\hline
\multirow{2}{*}{Methods} & \multicolumn{5}{c|}{Target domain} & \multicolumn{5}{c|}{Target domain}\\
\cline{2-11} 
 & \textit{photo} & \textit{art} & \textit{cartoon} & \textit{sketch} & avg. & \textit{photo} & \textit{art} & \textit{cartoon} & \textit{sketch} & avg. \\
\hline
 ERM & 16.27 & 16.62 & 18.40 & 12.01 & 15.82 & 43.29 & 24.27 & 32.62 & 20.84 & 30.26 \\
MoCo V2 & 44.19 & 25.85 & 35.53 & 24.97 & 32.64 & 59.86 & 28.58 & 48.89 & 34.79 & 43.03 \\
SimCLR V2 & 54.65 & 37.65 & 46.00 & 28.25 & 41.64 & 67.45 & 43.60 & 54.48 & 34.73 & 50.06 \\
BYOL & 27.01 & 25.94 & 20.98 & 19.69 & 23.40 & 41.42 & 23.73 & 30.02 & 18.78 & 28.49\\
AdCo & 46.51 & 30.31 & 31.45 & 22.96 & 32.81 & 58.59 & 29.81 & 50.19 & 30.45 & 42.26 \\
MAE & 35.89 & 25.59 & 33.28 & 32.39 & 31.79 & 36.84 & 25.24 & 32.25 & 34.45 & 32.20 \\
DARLING & 53.37 & 39.91 & 46.41 & 30.17 & 42.46 & 68.86 & 41.53 & 56.89 & 37.51 & 51.20 \\
DiMAE\tnote{*} \ \ & 77.87 & 59.77 & \second{57.72} & 39.25 & \second{58.65} & \second{78.99} & 63.23 & \second{59.44} & \second{55.89} & \second{64.39} \\
BrAD\tnote{*} \ \ & \second{72.17} & 44.20 & 50.01 & \second{55.66} & 55.51 & \xmark & \xmark & \xmark & \xmark & \xmark \\
 CycleMAE\tnote{*} \ \  & \first{85.94} & \first{67.93} & 59.34 & 38.25 & \first{62.87} & \first{90.72} & \first{75.34} & \first{69.33} & 50.24 & \first{71.41} \\
\hline
\hline
\rowcolor{apricot!15!} SimCLR w/ FA & 62.67 & 49.92 &	54.79 &	58.32 &	56.43 & 78.36 &	59.41 &	65.16 &	63.59 &	66.63 \\
\rowcolor{apricot!30!} SimCLR w/ BSS & 63.29 \up{0.62} & \second{51.37} \up{1.45} & \first{59.43} \up{4.64} & \first{66.09} \up{7.77} & 60.04 \up{3.61}  & 79.50 \up{1.14} & \second{62.73} \up{3.32} & 65.67 \up{0.51} & \first{73.02} \up{9.43} & 70.23 \up{3.6}\\
\hline
\rowcolor{apricot!15!} SWaV w/ FA & 46.27 & 44.68 & 50.27 & 50.02 & 47.81 & 77.50 & 57.49 & 64.32 & 66.08 & 66.35 \\
\rowcolor{apricot!30!} SWaV w/ BSS & 57.82 \up{11.55} & 45.91 \up{1.23} & 53.65 \up{3.38} & 55.67 \up{5.65} & 53.27 \up{5.46} & 78.62 \up{1.12} & 59.65 \up{2.16} & 65.40 \up{1.08} & 67.80 \up{1.72} & 67.87 \up{1.52} \\
\hline
\end{tabular}
}
\begin{tablenotes}[para]
   \item[*] Uses Imagenet transfer learning.
\end{tablenotes}
\end{threeparttable}
\end{table}
\endgroup

\textbf{DomainNet.} Following prior UDG methods, \textit{painting}, \textit{real} and \textit{sketch} were selected as source domains and others as target domains. The reversed domains combination was also considered. For these two combinations, for each fraction of labeled data (1\%, 5\% and 10\%) and each of our SSL models, we report on Table~\ref{tab:domainnet_udg_performances}, the accuracy on each target domain, the per-domain averaged accuracy and the overall accuracy. The presented results are averaged over $3$ independent runs.
\begingroup
\setlength{\tabcolsep}{3pt}
\begin{table}[ht]
\caption{UDG performances on DomainNet subset. 
Best methods are highlighted in \first{bold}.}
\label{tab:domainnet_udg_performances}
\centering
\resizebox{\linewidth}{!}{%
\begin{tabular}{|l|lll|lll|ll|}
\hline
\multicolumn{1}{|c|}{Sources} & \multicolumn{3}{c|}{\textit{painting} $\cup$ \textit{real} $\cup$ \textit{sketch}} & \multicolumn{3}{c|}{\textit{clipart} $\cup$ \textit{infograph} $\cup$ \textit{quickdraw}} & \multicolumn{2}{c|}{} \\
\hline
\multicolumn{1}{|c|}{Target} & \textit{clipart} & \textit{infograph} & \textit{quickdraw} & \textit{painting} & \textit{real} & \textit{sketch} & Overall & Avg. \\
\hline
\rowcolor{blue!3} \multicolumn{9}{|c|}{Label Fraction $1\%$}\\
\hline
ERM & 6.54 & 2.96 & 5.00 & 6.68 & 6.97 & 7.25 & 5.88 & 5.90 \\
BYOL & 6.21 & 3.48 & 4.27 & 5.00 & 8.47 & 4.42 & 5.61 & 5.31 \\
MoCo V2 & 18.85 & 10.57 & 6.32 & 11.38 & 14.97 & 15.28 & 12.12 & 12.90 \\
AdCo & 16.16 & 12.26 & 5.65 & 11.13 & 16.53 & 17.19 & 12.47 & 13.15 \\
SimCLR V2 & 23.51 & 15.42 & 5.29 & \second{20.25} & 17.84 & 18.85 & 15.46 & 16.86 \\
DARLING & 18.53 & 10.62 & 12.65 & 14.45 & 21.68 & 21.30 & 16.56 & 16.54 \\
DiMAE & 26.52 & 15.47 & 15.47 & 20.18 & \second{30.77} & 20.03 & 21.85 & 21.41 \\
BrAD & \second{47.26} & \second{16.89} & \second{23.74} & 20.03 & 25.08 & \second{31.67} & \second{25.85} & \second{27.45} \\
CycleMAE & 37.54 & 18.01 & 17.13 & 22.85 & 30.38 & 22.31 & 24.08 & 24.71 \\
\hline
\rowcolor{apricot!15!} SimCLR w/ FA & 60.83 & 18.42 & 26.31 & 24.29 & 29.73 & 40.29 & 30.82 & 33.31 \\
\rowcolor{apricot!30!} SimCLR w/ BSS & \first{61.94} \up{1.11} & 19.58 \up{1.16} & \first{26.98} \up{0.67} & 27.40 \up{3.11} & 31.55 \up{1.82} & 41.49 \up{1.2} & 32.27 \up{1.45} & 34.82 \up{1.51} \\
\hline
\rowcolor{apricot!15!} SWaV w/ FA & 59.27 & \first{20.95} & 18.94 & 30.99 & 35.73 & 45.28 & 31.87 & 35.19 \\
\rowcolor{apricot!30!} SWaV w/ BSS & 60.40 \up{1.13} & 20.12 \down{0.83} & 23.09 \up{4.15} & \first{34.64} \up{3.65} & \first{38.45} \up{2.72} & \first{46.90} \up{1.62} & \first{34.32} \up{2.45} & \first{37.27} \up{2.08} \\
\hline
\hline
\rowcolor{blue!3} \multicolumn{9}{|c|}{Label Fraction $5\%$}\\
\hline
ERM & 10.21 & 7.08 & 5.34 & 7.45 & 6.08 & 5.00 & 6.50 & 6.86 \\
BYOL & 9.60 & 5.09 & 6.02 & 9.78 & 10.73 & 3.97 & 7.83 & 7.53 \\
MoCo V2 & 28.13 & 13.79 & 9.67 & 20.80 & 24.91 & 21.44 & 18.99 & 19.79 \\
AdCo & 30.77 & 18.65 & 7.75 & 19.97 & 24.31 & 24.19 & 19.42 & 20.94 \\
SimCLR V2 & 34.03 & 17.17 & 10.88 & 21.35 & 24.34 & 27.46 & 20.89 & 22.54 \\
DARLING & 39.32 & 19.09 & 10.50 & 21.09 & 30.51 & 28.49 & 23.31 & 24.83 \\
DiMAE & 42.31 & 18.87 & 15.00 & 27.02 & \second{39.92} & 26.50 & 27.85 & 28.27 \\
BrAD & \second{64.01} & \first{25.02} & \second{29.64} & \second{29.32} & 34.95 & \second{44.09} & \second{35.37} & \second{37.84} \\
CycleMAE & 55.14 & 20.87 & 19.62 & 27.64 & 40.24 & 28.71 & 30.80 & 32.04 \\
\hline
\rowcolor{apricot!15!} SimCLR w/ FA & 69.04 & 20.31 & 29.76 & 36.44 & 41.95 & 51.05 & 38.60 & 41.42 \\
\rowcolor{apricot!30!} SimCLR w/ BSS & \first{71.21} \up{2.17} & 20.93 \up{0.62} & \first{32.42} \up{2.66} & 36.68 \up{0.24} & 41.49 \down{0.46} & 52.75 \up{1.7} & 39.73 \up{1.13} & 42.58 \up{1.16} \\
\hline
\rowcolor{apricot!15!} SWaV w/ FA & 68.84 & 24.05 & 26.06 & 43.97 & 49.11 & 59.16 & 41.68 & 45.20 \\
\rowcolor{apricot!30!} SWaV w/ BSS & 70.56 \up{1.72} & \second{24.35} \up{0.3} & 28.83 \up{2.77} & \first{46.17} \up{2.2} & \first{51.21} \up{2.1} & \first{59.71} \up{0.55} & \first{43.53} \up{1.85} & \first{46.81} \up{1.61} \\
\hline
\hline
\rowcolor{blue!3} \multicolumn{9}{|c|}{Label Fraction $10\%$}\\
\hline
ERM & 15.10 & 9.39 & 7.11 & 9.90 & 9.19 & 5.12 & 8.94 & 9.30 \\
BYOL & 14.55 & 8.71 & 5.95 & 9.50 & 10.38 & 4.45 & 8.69 & 8.92 \\
MoCo V2 & 32.46 & 18.54 & 8.05 & 25.35 & 29.91 & 23.71 & 21.87 & 23.00 \\
AdCo & 32.25 & 17.96 & 11.56 & 23.35 & 29.98 & 27.57 & 22.79 & 23.78 \\
SimCLR V2 & 37.11 & 19.87 & 12.33 & 24.01 & 30.17 & 31.58 & 24.28 & 25.84 \\
DARLING & 35.15 & 20.88 & 15.69 & 25.90 & 33.29 & 30.77 & 26.09 & 26.95 \\
DiMAE (full fine-tune) & \second{70.78} & \second{38.06} & 27.39 & \second{50.73} & \second{64.89} & 55.41 & 49.49 & 51.21 \\
BrAD & 68.27 & \second{26.60} & \first{34.03} & 31.08 & 38.48 & 48.17 & 38.74 & 41.10 \\
CycleMAE(full fine-tune) & \first{74.87} & \first{38.42} & 28.32 & \first{52.81} & \first{67.13} & 56.37 & \first{50.78} & \first{52.98} \\
\hline
\rowcolor{apricot!15!} SimCLR w/ FA & 70.12 & 20.50 & 31.23 & 39.16 & 44.45 & 52.87 & 40.31 & 43.05 \\
\rowcolor{apricot!30!} SimCLR w/ BSS & 71.95 \up{1.83} & 21.27 \up{0.77} & 33.47 \up{2.24} & 39.49 \up{0.33} & 44.67 \up{0.22} & 55.42 \up{2.55} & 41.57 \up{1.26} & 44.38 \up{1.33} \\
\hline
\rowcolor{apricot!15!} SWaV w/ FA & 69.81 & 24.39 & 28.97 & 45.92 & 50.79 & \first{60.78} & 43.46 & 46.78 \\
\rowcolor{apricot!30!} SWaV w/ BSS & 71.99 \up{2.18} & 24.34 \down{0.05} & 29.82 \up{0.85} & 48.28 \up{2.36} & 52.37 \up{1.58} & 60.55 \down{0.23} & 44.59 \up{1.13} & 47.89 \up{1.11} \\
\hline
\end{tabular}
}
\end{table}
\endgroup
When integrating BSS, almost all target domain accuracies increase while per-domain-averaged accuracy always improves: SimCLR $\rightarrow (+1.51\%, +1.16\%, +1.33\%)$; SWaV $\rightarrow (+2.08\%, +1.61\%, +1.11\%)$ for the labeled data fractions $1\%$, $5\%$ and $10\%$, respectively.

\textbf{Camelyon17 WILDS.}
Averaged accuracy over $10$ independent runs, for different fractions of labeled data ($1\%$, $5\%$, $10\%$, $100\%$), on the \texttt{test} split are reported on Table~\ref{tab:camelyon17_wilds_udg_performances}. To ensure fair comparisons, the reimplemented methods DARLING and DiMAE used identical pretraining and fine-tuning hyperparameters as extended SSL methods.
\begin{table}[ht]
\caption
{
UDG performances on Camelyon17 WILDS. 
Best methods are highlighted in \first{bold}.}
\label{tab:camelyon17_wilds_udg_performances}
\centering
\begin{threeparttable}[b]
\resizebox{0.60\columnwidth}{!}{%
\begin{tabular}{|l|llll|}
\hline
 & \multicolumn{4}{c|}{\cellcolor{blue!3}Label Fraction}\\
\hline
Method & 1\% & 5\% & 10\% & 100\%\\
\hline
DARLING\tnote{*} & 70.44 & 72.00 & 72.43 & 72.36 \\
DiMAE\tnote{*} & \second{89.81} & \second{89.17} & \second{89.77} & \second{90.40} \\
\hline
FixMatch\tnote{\textdagger} & \xmark & \xmark & \xmark & 71.00\\
SWaV\tnote{\textdagger} & \xmark & \xmark & \xmark & 91.40\\
\hline
\rowcolor{apricot!15!} SimCLR w/ FA & 90.82 & 93.00 & 92.94 & 93.44\\
\rowcolor{apricot!30!} SimCLR w/ BSS & \first{92.27} \up{1.45} & \first{94.75} \up{1.75} & \first{94.82} \up{1.88} & \first{95.00} \up{1.56} \\
\hline
\rowcolor{apricot!15!} SWaV w/ FA & 91.03 & 91.94 & 91.96 & 92.26\\
\rowcolor{apricot!30!} SWaV w/ BSS & \first{93.42} \up{2.39} & \first{93.99} \up{2.05} & \first{94.07} \up{2.11} & \first{94.08} \up{1.82} \\
\hline
\rowcolor{apricot!15!} MSN w/ FA & 87.07 & 87.33 & 87.55 & 88.82\\
\rowcolor{apricot!30!} MSN w/ BSS & \first{90.93} \up{3.86} & \first{91.82}  \up{4.49}& \first{91.83} \up{4.28} & \first{91.98} \up{3.76} \\
\hline
\end{tabular}
}
\begin{tablenotes}
   \item[*] Our implementation (no available public code).
   \item[\textdagger] From WILDS challenge~\citep{koh2021wilds}. Uses unlabeled data from the target
   domain.
\end{tablenotes}
\end{threeparttable}
\end{table}
For each extended SSL method, BSS leads to substantial performance gains ranging from $+1.45\%$ to $+4.49\%$ and results in state-of-the-art performances. Extended SSL methods with BSS even surpass FixMatch and SWaV trained with additional unlabeled data from the target domain.

\section{Ablation studies and additional experiments}
\subsection{Augmentation strategy ablation study}
To evaluate the benefits of each component in our augmentation strategy (BSS + additional augmentations) on performances, we pretrained a model on Camelyon17 WILDS using SimCLR and different combinations of components. For each combination and each fraction of labeled data, averaged accuracy over $10$ independent runs are reported in Table~\ref{tab:ablation_studies}.
\begin{table}[ht]
\caption{SimCLR's performances on Camelyon17 WILDS varying our augmentation strategy's components. Best methods are highlighted in \first{bold}.
}
\label{tab:ablation_studies}
\centering
\resizebox{0.6\columnwidth}{!}{%
\begin{tabular}{|c|c|c|cccc|}
\hline
 \multicolumn{3}{|c|}{} & \multicolumn{4}{c|}{\cellcolor{blue!3}Label Fraction}\\
\hline
Color-jitter & FA & BSS & 1\% & 5\% & 10\% & 100\% \\
\hline
 Sample-wise & \xmark & \xmark & 63.54 & 63.58 & 63.50 & 65.06 \\
 Batch-wise & \xmark & \xmark & 66.81 & 68.19 & 68.45 & 68.53 \\
 Sample-wise & \cmark & \xmark & 90.82 & 93.00 & 92.94 & 93.44 \\
 Batch-wise & \xmark & \cmark & \first{92.27} & \first{94.75} & \first{94.82} & \first{95.00} \\
\hline
\end{tabular}
}
\end{table}
Sample-wise color jittering and no FA or BSS, resulting in the regular SimCLR, leads to poor performance for the different fractions of labeled data. Changing the color-jittering from sample-wise to batch-wise slightly improves performances suggesting that even reducing the styles variability of augmented images helps for generalization. FA leads to drastic performance gains compared to the regular SimCLR which is not surprising given their prior success in DG tasks. Combining SimCLR with BSS and batch-wise color-jitter yields even greater performance improvements and non-negligible gains compared to FA. This observation is also supported by results from Tables~\ref{tab:pacs_udg_performances},~\ref{tab:domainnet_udg_performances},~\ref{tab:camelyon17_wilds_udg_performances}.  

\subsection{Underlying mechanisms involved in BSS efficiency}
\label{subsec:additional_experiments}
\begin{figure}[ht]
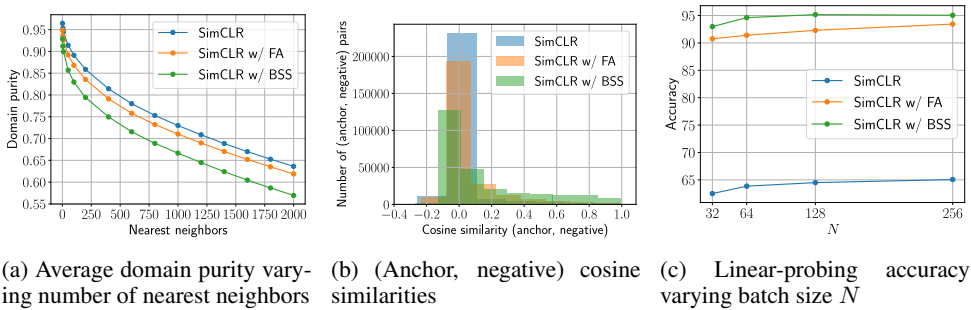

    \centering
    \begin{tcolorbox}[colframe=white, colback=white]
    \begin{subfigure}[t]{0.32\textwidth}
        \centering
        \includesvg[width=\linewidth]{imgs/simclr_domain_purities_new}
        \caption{Average domain purity varying number of nearest neighbors}
        \label{subfig:domain_purities}
    \end{subfigure}
    \hfill
    \begin{subfigure}[t]{0.32\textwidth}
        \centering
        \includesvg[width=\linewidth]{imgs/distances_anchor_negatives_new}
        \caption{(Anchor, negative) cosine similarities}
        \label{subfig:anchor_negative_similarities}
    \end{subfigure}
    \hfill
    \begin{subfigure}[t]{0.32\textwidth}
        \centering
        \includesvg[width=\linewidth]{imgs/fourier_based_contrastive_learning_batch_sizes_experiment_new}
        \caption{Linear-probing accuracy varying batch size $N$}
        \label{subfig:simclr_bss_batch_dependency}
    \end{subfigure}
    \end{tcolorbox}
    \caption{\update{SimCLR experiments with standard augmentation, FA or BSS on Camelyon17 WILDS.}}
    \label{fig:simclr_extended_experiments}
\end{figure}

\textbf{Spurious correlations reduction, harder negatives creation and reduced batch size requirement (SimCLR). }
We hypothesized that BSS should help reduce the emergence of spurious correlations when repelling negatives from positives. Additionally, standardizing styles among positives and negatives should also facilitate the creation of harder negatives, a known factor contributing to robust performance~\citep{kalantidis2020hard, robinson2020contrastive}, while also reducing the demand for large batch sizes. To validate these hypotheses, we conducted $3$ comprehensive experiments employing \update{SimCLR with standard augmentation, FA, or BSS on Camelyon17 WILDS}: 
(1) To validate BSS's effectiveness in reducing spurious correlations, we computed the averaged domain purity for representations of unseen source and target examples after pretraining. This metric, which quantifies the degree to which each example and its nearest neighbors share the same domain label, serves as an indicator of the domain-invariance within SSL representations (refer to Figure~\ref{subfig:domain_purities}).
(2) To assess the impact of BSS on encouraging the presence of harder negatives, we computed representations for several augmented batches and calculated cosine similarities across all possible (anchor, negative) pairs, reporting the values in a histogram (refer to Figure~\ref{subfig:anchor_negative_similarities}).
(3) Lastly, to assess BSS's ability to mitigate the demand for large batch sizes, we pretrained SimCLR varying batch sizes and assessed performance using linear probing (refer to Figure~\ref{subfig:simclr_bss_batch_dependency}).
\update{Figure~\ref{subfig:domain_purities} demonstrates that SimCLR with FA exhibits slightly lower average domain purity than standard SimCLR while SimCLR with BSS results in much lower average domain purity. This observation affirms BSS's effectiveness in attenuating spurious correlations and enhancing domain-invariance. Figure~\ref{subfig:anchor_negative_similarities} indicates that standard SimCLR tends to produce negatives that are dissimilar to the anchors. SimCLR with FA produces negatives more similar to the anchors but not to the same extent as SimCLR with BSS which confirms BSS's role in creating harder negatives.
Finally, Figure~\ref{subfig:simclr_bss_batch_dependency} reveals that standard SimCLR yields considerably lower performances compared to SimCLR with FA or with BSS. FA and BSS lead to improved performances for any batch size. As batch size increases, performances augment until a plateau is reached. However, when using BSS, this plateau is reached for a lower batch size supporting BSS’s efficacy in reducing the need for large batches.}
\par
\begin{figure}[ht]
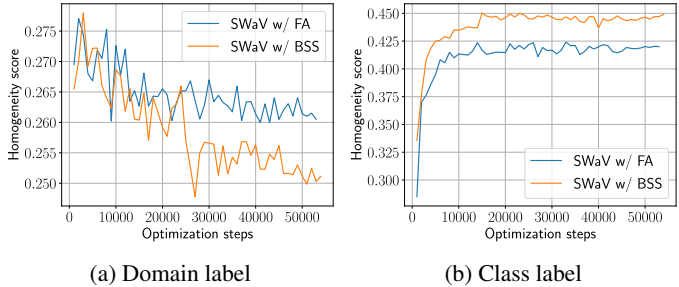

    \centering
    \begin{subfigure}[t]{0.32\textwidth}
        \centering
        \includesvg[width=\linewidth]{imgs/domain_homogeneity_experiments}
        \caption{Domain label}
        \label{subfig:assignements_domain_homogeneity}
    \end{subfigure}
    \begin{subfigure}[t]{0.32\textwidth}
        \centering
        \includesvg[width=\linewidth]{imgs/label_homogeneity_experiments}
        \caption{Class label}
        \label{subfig:assignements_label_homogeneity}
    \end{subfigure}
    \caption{Averaged homogeneity scores for representations assigned to the same prototype.}
    \label{fig:}
\end{figure}

\textbf{Better domain heterogeneity and class homogeneity for examples assigned to the same prototype (SWaV, MSN).}
We postulated that the coexistence of multiple domains/styles within views used for cluster assignments computation (i.e.: global views for SWaV and unmasked views for MSN) could introduce correlations between the assignments and domains/styles. To investigate the effectiveness of BSS in mitigating these correlations and shed light on why BSS yields superior representations compared to FA, we conducted the following experiment: We pretrained a backbone with SWaV using FA or BSS on DomainNet (sources:~$\textit{painting} \cup \textit{real} \cup \textit{sketch}$, targets:~$\textit{clipart} \cup \textit{infograph} \cup \textit{quickdraw}$). During training, at every $1K$ optimization step, we computed representations from unseen source and target examples along with their hard assignments resulting from SK. Subsequently, we evaluated the homogeneity of representations assigned to each prototype in terms of domain or class labels and averaged the homogeneity scores over all prototypes. The evolution of the averaged homogeneity score with respect to domain or class labels are respectively reported in Figure~\ref{subfig:assignements_domain_homogeneity} and Figure~\ref{subfig:assignements_label_homogeneity}. 
Figure~\ref{subfig:assignements_domain_homogeneity} reveals that employing BSS instead of FA tends to reduce domain homogeneity (or improve domain heterogeneity) among representations assigned to the same prototype. This observation confirms BSS's role in reducing correlations between assignments and domains. Conversely, Figure~\ref{subfig:assignements_label_homogeneity} illustrates that BSS results in higher class label homogeneity attesting that BSS helps to produce assignments more semantically coherent.
\section{Conclusion}
This work introduces Batch Styles Standardization, an image style standardization technique to be combined with existing SSL methods to address UDG. Extending existing SSL methods with BSS offers serious advantages over prior UDG methods, including the elimination of domain labels and domain-specific network components dependencies to enhance domain-invariance while offering versatility for integration. Leveraging BSS, the extended SSL methods exhibit improved generalization capabilities, often surpassing or competing with alternative UDG strategies. Comprehensive experiments provide insights into the underlying mechanisms involved in BSS's efficiency. Finally, other style transfer techniques, whether at the image-level (such as GAN-based methods) or features-level (e.g., AdaIN), may be used to standardize images style to mitigate spurious correlations in SSL methods. However, we leave the exploration of these possibilities for future research.

\clearpage


\appendix
\section{Pseudo-code and PyTorch implementation of Batch Styles Standardization}
\label{app:pseudo_code}
On Algorithm~\ref{alg:BSS} and Listing~\ref{lst:bbs_PyTorch_implementation}, a pseudo-code along with a PyTorch implementation of Batch Styles Standardization are provided. The full code will be released upon acceptance.
\begin{algorithm}[ht]
    \SetKwFunction{substitutelowfreq}{substitute\_low\_freq}
    \SetKwInOut{KwIn}{Input}
    \SetKwInOut{KwOut}{Output}

    \KwIn{
    \begin{itemize}
        \item $\{\mX_i\}_{1 \leq i \leq N}$: Batch of images\\
        \item $(r_{min}, r_{max})$: Minimum and maximum area ratios between the substituted amplitudes components and the full amplitude.
    \end{itemize}
    }
    \KwOut{
    Batch of images with standardized style: $\{\hat{\mX}_i\}_{1 \leq i \leq N}$
    }
    \tcp{Computes Fourier transform for all images in the batch}
    \For{$i \leftarrow 1$ \KwTo $N$}{
        $\mathcal{A}(\mX_i) \leftarrow \sqrt{  \operatorname{Re}\left (\mathcal{F}(\mX_i)\right ) ^2 + \operatorname{Im}\left (\mathcal{F}(\mX_i)\right ) ^2}$
        \\
        $\mathcal{P}(\mX_i) \leftarrow \arctan \left(
                \dfrac{\operatorname{Im}\left (\mathcal{F}(\mX_i)\right )}{\operatorname{Re}\left (\mathcal{F}(\mX_i)\right )}
            \right)$
    }
    \tcp{Sample the index $k$ associated to the style image and sample the amplitudes area ratio $r$}
    $k \sim U(\{1, \cdots, N\}) $ \\
    $r \sim U(r_{min}, r_{max}) $ \\
    \tcp{For each image, substitute the low-frequency components with those of the style image then apply the inverse Fourier transform to transfer the style onto the original image.}
    \For{$i \leftarrow 1$ \KwTo $N$}{
        $\hat{\mathcal{A}}(\mX_i) \leftarrow \substitutelowfreq(\mathcal{A}(\mX_i), \mathcal{A}(\mX_k), r)$\\
        $\hat{\mX}_i \leftarrow \mathcal{F}^{-1}\left (\hat{\mathcal{A}}(\mX_i)e^{-i\mathcal{P}(\mX_i)}\right)$
        
    }
    \KwRet{$\{\hat{\mX}_i\}_{1 \leq i \leq N}$}
    \caption{Batch Styles Standardization}
    \label{alg:BSS}
\end{algorithm}

\begin{listing*}[ht]
\begin{multicols}{2}
\begin{minted}[
    % frame=lines,
    fontsize=\tiny]{python}
class BatchStylesStandardization():
    """Implements Batch Styles Standardization. Given a
        batch of N images and their Fourier transforms,
        we manipulate the different amplitudes by
        substituting their low-frequency components
        with those a randomly chosen image.

    Attributes:
        ratios (tuple): $(r_min, r_max)$ specifying
            the minimum and maximum possible
            areas ratio between the substituted
            amplitude and the full amplitude.
    """
    def __init__(self, ratios):
        self.ratios = ratios

    def substitute_low_freq(self, src_amp, tgt_amp, ratio):
        """Substitute the low-frequency components
            of the source amplitudes with those
            of the target amplitudes. 

        Args:
            src_amp (torch.Tensor): Source amplitudes
            tgt_amp (torch.Tensor): Target amplitudes
            ratio (float): Area ratio between the
                substituted amplitude and the full
                amplitude.

        Returns:
            torch.Tensor: Source amplitudes where 
                the low-frequency components have been
                substituted with those
                of the target amplitudes.
        """
        # Compute center coordinates of amplitudes
        h, w = src_amp.shape[-2:]
        hc, wc = int(h//2), int(w//2)

        # Compute half length `l` of the components 
        # to be substituted
        l = min([int(ratio*h/2), int(ratio*w/2)])

        # Substitute low freq components of source
        # amplitudes with those of the target amplitudes
        low_freq_tgt_amp = tgt_amp[
            ..., hc-l:hc+l, wc-l:wc+l]
        src_amp[
            ..., hc-l:hc+l, wc-l:wc+l] = low_freq_tgt_amp
        return src_amp


    def __call__(self, imgs, n_views):
        """Apply batch styles standardization `n_views` times 
            on a batch of $N$ images.
            
        Args:
            imgs (torch.Tensor): Batch of images (N, 3, H, W)
            n_views (int): Number of augmented views

        Returns:
            torch.Tensor: Batch with standardized styles 
            (N, n_views, 3, H, W)
        """
        # Apply FFT on source images
        fft = torch.fft.fftn(
            imgs, dim=(-2, -1))
        # Shift low-frequency components to the center
        fft = torch.fft.fftshift(
            fft, dim=(-2, -1))
        # Retrieve amplitude and phase
        amp, phase = fft.abs(), fft.angle()

        # Sample n_views images that will be used as
        # ref styles
        bs = imgs.size(0)
        sampled_ind = torch.randperm(bs)[:n_views]

        # Substitute low-freq of src amplitudes with those
        # of the n_views sampled images
        src_amp = amp.unsqueeze(1).repeat(
            [1, n_views, 1, 1, 1])
        tgt_amp = amp[sampled_ind].unsqueeze(0).expand(
            bs, -1, -1, -1, -1)
        sampled_ratio = random.uniform(*self.ratios)
        amp = self.substitute_low_freq(
            src_amp, tgt_amp, sampled_ratio)

        phase = phase.unsqueeze(1)
        # Reconstruct FFT from amp and phase
        fft = torch.polar(amp, phase)
        # Shift back low-frequency to their
        # original positions
        fft = torch.fft.ifftshift(fft, dim=(-2, -1))
        # Invert FFT
        imgs = torch.fft.ifftn(
            fft, dim=(-2, -1)).real.clamp(0, 1)
        return imgs
\end{minted}
\end{multicols}
\caption{Batch Styles Standardization PyTorch implementation}
\label{lst:bbs_PyTorch_implementation}
\end{listing*}

\clearpage
\section{Technical details about SSL methods}
\label{app:ssl_methods_technical_details}

\subsection{SimCLR}
SimCLR aims to bring representations of augmented views of the same image closer (positives) while repelling all other images representations (negatives). 
In practice, given a batch of $N$ images, each image is augmented $V$ times independently resulting in a $N \times V$ images grid where each row $c$ corresponds to a content and each column $s$ to a view. For each image $\mX_{cs}$ and its corresponding representation $\vz_{cs} \in \R^{D}$, SimCLR minimizes the NT-Xent loss with temperature $T$:
\begin{equation}
    \label{eq:simclr_loss}
    \mathcal{L}_{\red{c}\blue{s}} =  \displaystyle \dfrac{- 1}{V-1} \sum_{s' \neq \blue{s}} \log \left (
    \dfrac{e^ { \vz_{\red{c}\blue{s}} \cdot \vz_{\red{c}s'} / T}}{
        \displaystyle \sum_{\tiny (c'',s'') \neq (\red{c},\blue{s})} e^{\vz_{\red{c}\blue{s}} \cdot \vz_{c''s''} / T}
    } \right)
\end{equation}

\subsection{SWaV}
SWaV computes the representations of different views of the same image while clustering them using an online algorithm. Since representations should capture similar information, SWaV assumes that one view's cluster assignment can predicted from representations of other views. This swapped prediction idea is the core concept behind SWaV loss formulation.

Concretely, in SWaV, an image $\mX_n$ is augmented into $2$ views $\mX_n^{(s)}$ and $\mX_n^{(t)}$, with corresponding representations $\vz_n^{(s)}$ and $\vz_n^{(t)}$. Similarities between representations and $K$ learnable cluster centroids/prototypes $\mC \in \R^{K \times D}$ are computed and converted into probabilities such as follows:
\begin{align}
    \vp_n^{(v)} &= \text{softmax}\left(\dfrac{\vz_{n}^{(v)} \cdot \mC^T}{\tau} \right) \ \ \forall v \in \{s, t\} 
\end{align}
To compute cluster assignments also referred to as codes and denoted $\vq_n^{(v)}$, SWaV relies on the Sinkhorn-Klopp (SK) algorithm~\citep{cuturi2013sinkhorn}. SK is performed on all views representations trying to assign representations to the most similar centroids but also uniformly among clusters. Finally, based on the swapped prediction concept, SWaV minimizes the following per-sample loss:
\begin{equation}
    \label{eq:swav_no_multi_crop_loss}
    \mathcal{L}_n = H(\vq_n^{(s)}, \vp_n^{(t)}) + H(\vq_n^{(t)}, \vp_n^{(s)}) 
\end{equation}
In Equation~\ref{eq:swav_no_multi_crop_loss}, $H(\vp, \vq)$ stands for the cross-entropy between an approximated probability distribution $\vq$ and a true probability distribution $\vq$.

In practice, SWaV employs a multi-crop strategy, generating $2$ global views (large crops) and $V$ local views (small crops) for each image. Cluster assignments are then computed only from the $2$ global views while probabilities are derived from all the $V+2$ views. In this setting, SWaV minimizes the following loss:
 \begin{equation}
    \mathcal{L}_n = \dfrac{1}{2(V+1)}\displaystyle \sum_{i=1}^{2}\sum_{v = 1}^{V+2} \1_{i \neq v} H(\vq_n^{(i)}, \vp_n^{(v)}) 
\end{equation}

\subsection{MSN}
Given two views of the same image, MSN randomly masks the patches of one view and leaves the other unchanged. Then, MSN's goal is to match the representation of the masked view with that of the unmasked view. 

To derive a view's representation, MSN computes the similarities between its embedding and a set of cluster centroids/prototypes, subsequently transforming them into a probability distribution. As direct matching of these representations can lead to representation collapse, MSN simultaneously optimizes a cross-entropy term along with an entropy regularization term on the mean representation of the masked views. The entropy regularization term encourages the model to use the entire set of centroids/prototypes. Additionally, MSN employs Sinkhorn-Klopp on the representations of the unmasked views to avoid tuning the hyperparameter weighting the entropy regularization term.

In practice and more formally, MSN generates for each image $\mX_n$, $M$ masked views $\{\mX_{n, 1}, \hdots, \mX_{n, M}\}$ and a single unmasked view $\mX_{n}^+$. Masked views are processed by a student encoder and the unmasked view by a teacher encoder whose weights are updated via an exponential moving average of the student encoder's weights. Masked and unmasked views' embeddings denoted $\{\vz_{n, 1}, \hdots, \vz_{n, M}\}, \vz_n^+$ are then compared to a set of centroids/prototypes $\mC \in \R^{K \times D}$ and the resulting similarities are converted into probability distributions $\{\vp_{n, 1}, \hdots, \vp_{n, M}\}, \vp_n^+$:
\begin{equation}
    \begin{cases}
    \vp_{n,m} = \text{softmax}\left(\dfrac{\vz_{n,m} \cdot \mC^T}{\tau}\right) \\
    \vp_{n}^+ = \text{softmax}\left(\dfrac{\vz_{n}^+ \cdot \mC^T}{\tau^+}\right)
    \end{cases}
\end{equation}
$\tau$ and $\tau^+$ stand for temperature hyperparameters and are chosen such that $\tau > \tau^+$ to encourage sharper probability distributions implicitly guiding the model to produce confident masked views representations. Given a batch of $N$ images, MSN minimizes the following loss:
\begin{equation}
    \begin{cases}
    \mathcal{L} = \dfrac{1}{NM} \displaystyle \sum_{n=1}^{N}\sum_{m=1}^{M} H(\vp_n^+, \vp_{n, m}) - \lambda H(\bar{\vp}) \\
    \bar{\vp} = \dfrac{1}{NM}\displaystyle \sum_{n=1}^{N}\sum_{m=1}^{M}\vp_{n, m}
    \end{cases}
\end{equation}
$H(\vp, \vq)$ stands for the cross-entropy between an approximated probability distribution $\vq$ and a true probability distribution $\vq$ while $H(\bar{\vp})$ denotes the entropy of the masked views' mean representation $\bar{\vp}$.

\section{Implementation details}
\label{app:implementation_details}
\subsection{Pretraining}
On \textbf{PACS} and \textbf{DomainNet}, as part of the geometric augmentations, we use random crop resizing, horizontal flips, small rotations, cutout\citep{devries2017improved} while color augmentations are applied in batch-wise manner using color jitter, random equalize, random posterize, random solarize and random grayscale.
On \textbf{Camelyon17 WILDS}, we use random crop resizing, flips, rotations, cutout and batch-wise color jitter.
All other hyperparameters for SimCLR, SWaV, MSN are respectively specified on Tables~\ref{tab:simclr_imp_details}, ~\ref{tab:swav_imp_details}, ~\ref{tab:msn_imp_details}.
\begin{table}[ht]
    \caption{Hyperparameters used for SimCLR extension based on Batch Styles Standardization}
    \label{tab:simclr_imp_details}
    \begin{center}
    \resizebox{\columnwidth}{!}{%
    \begin{tabular}{|c|c|c|c|}
         \hline
         datasets & PACS & DomainNet & Camelyon17 WILDS \\
         \hline
         backbone & ResNet-18 & ResNet-18 & ResNet-50 \\
         $V$ & $2 \times 224^2 + 6 \times 128^2$  & $2 \times 224^2 + 6 \times 128^2$ & $8 \times 128^2$ \\
         $(r_{min}, r_{max})$ & $(0.02, 1)$& $(0.02, 1)$ & $(0.02, 0.1)$ \\
         $D$ & $128$  & $128$ & $128$ \\
         $T$ & $0.5$& $0.5$ & $0.5$ \\
         $N$ & $256$& $512$ & $256$ \\
         steps & $60$K & $60$K & $150$K \\
         optimizer & LARS& LARS & LARS \\
         learning rate & $0.2$& $0.4$ & $0.2$ \\
         learning rate schedule &  linear warmp-up + cosine decay &  linear warmp-up + cosine decay &  linear warmp-up + cosine decay \\
         weight decay & $10^{-6}$& $10^{-6}$ & $10^{-6}$ \\
         \hline
    \end{tabular}
    }
    \end{center}
\end{table}
\begin{table}[ht]
    \caption{Hyperparameters used for SWaV extension based on Batch Styles Standardization}
    \label{tab:swav_imp_details}
    \begin{center}
    \resizebox{\columnwidth}{!}{%
    \begin{tabular}{|c|c|c|c|}
         \hline
         datasets & PACS & DomainNet & Camelyon17 WILDS \\
         \hline
         backbone & ResNet-18 & ResNet-18 & ResNet-50 \\
         global views & $2 \times 224^2$  & $2 \times 224^2$ & $2 \times 128^2$ \\
         local views & $6 \times 128^2$  & $6 \times 128^2$ & $6 \times 128^2$ \\
         $(r_{min}, r_{max})$ & $(0.02, 1)$& $(0.02, 1)$ & $(0.02, 0.1)$ \\
         $K$ & $256$  & $256$ & $256$ \\
         $D$ & $128$  & $128$ & $128$ \\
         $\tau$ & $0.1$& $0.1$ & $0.1$ \\
         $N$ & $256$& $512$ & $256$ \\
         steps & $60$K & $60$K & $150$K \\
         optimizer & LARS & LARS & LARS \\
         learning rate & $0.2$& $0.4$ & $0.2$ \\
         learning rate schedule &  linear warmp-up + cosine decay &  linear warmp-up + cosine decay &  linear warmp-up + cosine decay \\
         weight decay & $10^{-6}$& $10^{-6}$ & $10^{-6}$ \\
         \hline
    \end{tabular}
    }
    \end{center}
\end{table}
\begin{table}[ht]
    \caption{Hyperparameters used for MSN extension based on Batch Styles Standardization}
    \label{tab:msn_imp_details}
    \begin{center}
    \resizebox{0.5\columnwidth}{!}{%
    \begin{tabular}{|c|c|}
         \hline
         backbone & ViT-S/8\\
         unmasked views & $2 \times 96^2$   \\
         masked views &  $10 \times 64^2$ \\
         $(r_{min}, r_{max})$ & $(0.02, 0.1)$ \\
         patch masking ratio & $0.3$\\
         $K$ & $128$  \\
         $D$ & $384$  \\
         $\tau$ & $0.1$\\
         $\tau^+$ & $0.025$\\
         $N$ & $256$\\
         steps & $150$K\\
         Optimizer & LARS\\
         learning rate & $0.2$\\
         learning rate schedule & linear warm-up + cosine decay\\
         weight decay & $10^{-6}$\\
         EMA momentum & $0.995$\\
         \hline
    \end{tabular}
    }
    \end{center}
\end{table}

\subsection{fine-tuning/Linear-probing}
For all datasets (\textbf{PACS}, \textbf{DomainNet}, and \textbf{Camelyon17 WILDS}), we use the Adam optimization method~\citep{kingma2014adam} with an initial learning rate of $10^{-4}$, a learning rate scheduler with cosine decay, and weight decay of $10^{-4}$.
The networks are trained respectively for $5$K, $1$K, and $15$K steps with batch sizes of $128$, $64$, and $64$. When performing linear probing, we follow the same normalization scheme as~\citep{he2022masked} by adding a batch normalization layer~\citep{ioffe2015batch} without affine parameters before the linear classifier.

\section{Discussions \& additional visualizations}
\subsection{Transfer learning in DG/UDG}
\label{app:transfer_learning_dg_udg}
The usage of Transfer Learning in DG/UDG is common but we think it is misguided. Often the pretraining dataset, such as Imagenet, can include one or more of the target domains, \eg, \textit{photo} for PACS or \textit{real} for DomainNet. When evaluating on these domains, it is not possible to know if performances result from the generalization ability of the DG/UDG methods or from the transfer learning. For new DG/UDG methods, it is hard not to follow the common practice because transfer learning unfairly boosts the results of previous works, and state-of-the-art performances are often seen as a prerequisite for paper acceptance. We tried to limit the usage of transfer learning in our experiments and only used it for the DomainNet dataset. 
The bias introduced by Imagenet transfer learning can be seen in our results and especially for the domains that are close to Imagenet: on PACS (Table 1 main paper), without transfer learning, SimCLR with BSS is consistently the best method for all settings except for the \textit{photo} domain, where methods using Imagenet transfer learning report better performance. However, on DomainNet (Table 2 main paper), when using Imagenet transfer learning similar to the other UDG methods, SimCLR with BSS achieves better performances on the \textit{real} domain for 2 out of 3 proportions of labeled data ($1\%$, $5\%$). In the $10\%$ labeled data setting, where SimCLR with BSS is outperformed, it is worth noting that DiMAE is the only method to perform full fine-tuning which is probably the reason behind the gap in performances.   

\subsection{Features visualization}
\begin{figure}[ht]
    \begin{center}
    \includegraphics[width=\textwidth]{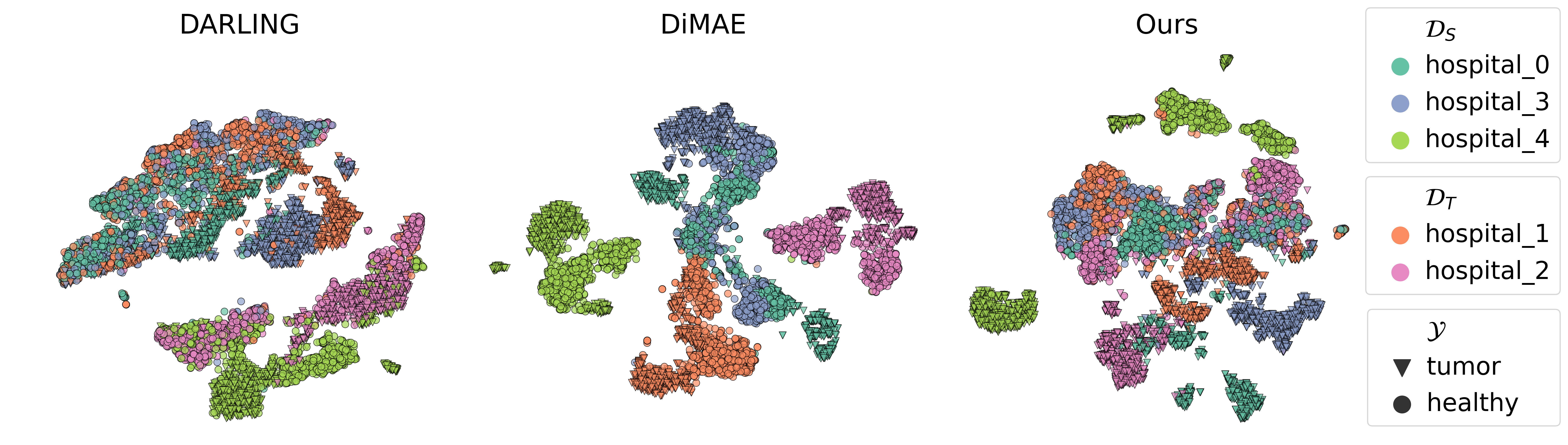}
    \end{center}
    \caption{t-SNE plots of the backbone representations for different UDG methods on Camelyon17 WILDS. Colors and markers correspond respectively to different domains and classes. On the target domains (\texttt{hospital\_1}, \texttt{hospital\_2}), our method (SimCLR w/ BSS) shows better domain confusion while keeping better class separability. Zoom on pdf for better visualization.}
    \label{fig:tsne_camelyon_17}
\end{figure}

To assess the quality of the SSL representations and their ability to generalize across domains, we display, in Figure~\ref{fig:tsne_camelyon_17}, t-SNE plots of the backbone representations for SimCLR with BSS and competitors on Camelyon17 WILDS.
DARLING representations tend to be domain-invariant as lots of examples from different domains are superimposed. However, this is also the case for many examples from different classes indicating potentially poor model generalization. In contrast, DiMAE representations appear to be well separated by classes but also by domains, especially for the target domains \texttt{hospital\_1} and \texttt{hospital\_2}, indicating a lack of domain-invariance. Finally, better class separability and domain confusion emerge from the representations of SimCLR with BSS revealing a better domain-invariance and a potentially better cross-domain generalization.
\subsection{Impact of \texorpdfstring{$r$}{Lg} on BSS generated images}
\begin{figure}[ht]
    \centering
    \includegraphics[width=\textwidth]{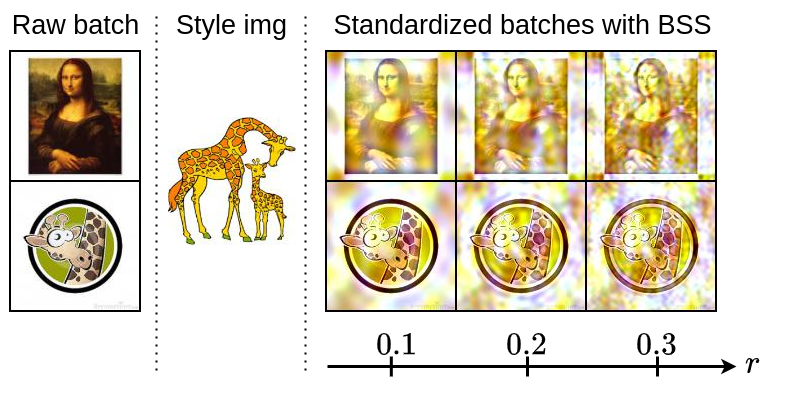}
    \caption{Impact of hyperparameter $r$ on augmented images with BSS}
    \label{fig:impact_of_ratio_hyperparameter}
\end{figure}

To illustrate the effect of the hyperparameter $r$ on the generated images by BSS, we apply BSS on a single batch fixing the chosen style image and varying $r$. The resulting images are reported in Figure~\ref{fig:impact_of_ratio_hyperparameter}. We can observe that as $r$ increases, textures/styles with higher frequencies are transferred to the resulting images. 

\section{Acknowledgements}
This project was provided with computer and storage resources by GENCI at IDRIS thanks to the grant 2022-AD011013424R1 on the supercomputer Jean Zay's the V100 partition.

\clearpage
{
\small
\bibliographystyle{iclr2024_conference}


\bibliography{iclr2024_conference}
}

\end{document}